\documentclass{article}

\usepackage{PRIMEarxiv}

\usepackage[utf8]{inputenc} 
\usepackage[T1]{fontenc}    
\usepackage{hyperref}       
\usepackage{url}            
\usepackage{booktabs}       
\usepackage{amsfonts}       
\usepackage{amsmath}
\usepackage{nicefrac}       
\usepackage{microtype}      
\usepackage{lipsum}
\usepackage{fancyhdr}       
\usepackage{graphicx}       
\graphicspath{{media/}}     

\title{Advancements in Upper Body Exoskeleton: Implementing Active Gravity Compensation with a Feedforward Controller
 \thanks{This work is supported by the Ministry of Economics, Innovation, Digitization and Energy of the State of North Rhine-Westphalia and the European Union, grants GE-2-2-023A (REXO) and IT-2-2-023 (VAFES).}
 \thanks{Muhammad Ayaz Hussain is with the Iossifidis-Lab, Institute of Computer Science, Ruhr West University of Applied Sciences, 45479 Mülheim an der Ruhr, Germany {\tt\small mayazhussain@gmail.com}}
 \thanks{Ioannis Iossifidis is head of the Iossifidis-Lab, Institute of Computer Science, Ruhr West University of Applied Sciences, 45479 Mülheim an der Ruhr, Germany}
}

\author{
  Muhammad Ayaz Hussain and Ioannis Iossifidis\\
  Institute of Computer Science\\
  Ruhr West University of Applied Science\\
  Mülheim an der Ruhr, Germany \\
  \texttt{\{Muhammad.Hussain, iossifidis\}@hs-ruhrwest.de} \\
}

\begin{document}

\maketitle
\thispagestyle{empty}
\pagestyle{empty}

\begin{abstract}
In this study, we present a feedforward control system designed for active gravity compensation on an upper body exoskeleton. The system utilizes only positional data from internal motor sensors to calculate torque, employing analytical control equations based on Newton-Euler Inverse Dynamics. Compared to feedback control systems, the feedforward approach offers several advantages. It eliminates the need for external torque sensors, resulting in reduced hardware complexity and weight. Moreover, the feedforward control exhibits a more proactive response, leading to enhanced performance.
The exoskeleton used in the experiments is lightweight and comprises 4 Degrees of Freedom, closely mimicking human upper body kinematics and three-dimensional range of motion. We conducted tests on both hardware and simulations of the exoskeleton, demonstrating stable performance. The system maintained its position over an extended period, exhibiting minimal friction and avoiding undesired slewing. 
\end{abstract}
\section{Introduction}
 In recent years, exoskeleton technology has emerged as a promising avenue for enhancing human mobility and assisting individuals with mobility impairments. These wearable robotic systems have shown great potential in various applications, ranging from augmenting strength and endurance for industrial workers to aiding patients with neurological disorders in regaining mobility. Among the key challenges the exoskeleton developers face is the effective management of gravity-related forces, which can be especially demanding in upper-body exoskeletons due to their complex kinematics and range of motion. As the designed exoskeleton is aimed towards rehabilitating patients, it has to be lightweight, and the wearer should experience the motion smoothly and with minimal muscle effort.  \newline
 To address this challenge, the present study focuses on the development of an innovative feedforward control system for active gravity compensation on an upper body exoskeleton. Since the reference is time-varying and is not known apriori, the torque has to be computed at every instance of time, therefore a Feedforward controller is chosen. Unlike traditional feedback, torque control systems that rely on external sensors and their corrective actions based on feedback errors can introduce jerk and integral windups. The feedforward approach aims to predict and counteract gravitational forces using only the positional data obtained from the motor's internal sensors.  This unique feature significantly reduces hardware complexity and weight, paving the way for more practical and efficient exoskeleton designs. \newline
 The proposed control system leverages analytical control equations based on Newton-Euler Inverse Dynamics, enabling real-time calculation of the corresponding torque to counteract gravity. By avoiding the need for external torque sensors, the system achieves a more proactive and responsive control, leading to improved performance and stability during operation. \newline
 The upper body exoskeleton at the core of this study boasts a lightweight and ergonomic design, mimicking human upper body kinematics and providing a three-dimensional range of motion. The system's four Degrees of Freedom offer versatility and adaptability, making it suitable for various activities and applications. \newline
 To evaluate the effectiveness of the feedforward control system, comprehensive testing was conducted on both hardware and simulations of the designed exoskeleton. The system's ability to maintain its position for an extended period, even in the presence of minimal friction, was observed, demonstrating the efficacy and robustness of the developed control mechanism which is discussed in the Section \ref{Results}  . \newline
 Overall, the implementation of a feedforward control system for active gravity compensation on an upper body exoskeleton holds great promise in advancing the field of wearable robotics as it can be implemented on most manipulator arms as they have position sensors by default. Its potential to enhance human-machine interactions and its application in rehabilitation and assistive technologies present exciting prospects for addressing real-world mobility challenges and improving the quality of life for a diverse range of users.

\section{Related Works}
The gravity compensation has been studied and implemented on several rehabilitative exoskeletons, both upper-body and lower-body, and both in active and passive domains.
The authors of  \cite{One-G} coined the terms Zero-G and One-G in the context of the upper body exoskeleton (CyberForce Exoskeleton), where One-G meant that only the weight of the rehabilitative exoskeleton is compensated, whereas Zero-G meant that both the weight of the exoskeleton, and the weight of the patient’s arm, are compensated. They calculated the amount of force exerted by the robot using haptic feedback and discretizing the workspace into cubes and calculated force at each of their vertices. Afterward, the authors performed Electromyography to study muscle activity and fatigue in both Zero-G and One-G scenarios in virtual environments, and it was found out that for Zero-G scenarios, the muscle fatigue was less than One-G scenario and almost similar to the fatigue when the subject is resting. In our paper, we used the same terms for Zero-G and One-G as used by the authors of this paper. \\
In \cite{Freebal}, the authors looked into the improvement in the work area of the hemiparetic arm by performing passive gravity compensated (using Freebal device) reach training on seven chronic stroke patients for 6 weeks in 18 half-hour sessions and they concluded the mean increase in Fugl-Meyer Assessment by 3 points, which indicates the improvement of the motor functioning, balance, and joint functioning in patients suffering from post-stroke hemiplegia by the means of gravity compensation.\\
In \cite{Harmony1}, the authors discussed the modeling, control, and kinematic and design evaluation of the HARMONY exoskeleton. The control system was based on a recursive Newton-Euler algorithm, which was formulated as a part of the dynamic model of the robot based on the feed-forward torque, which compensates for the robot dynamics. The exoskeleton has five active degrees of freedom and one degree of freedom each for elbow and wrist joints using Series Elastic Actuators. The feedback of those actuators was utilized by a PID controller to control the torque output.\\
The authors of \cite{Harmony1} in \cite{Harmony2} presented a strategy for the control of the shoulder mechanism of an upper-body exoskeleton (HARMONY) for promoting the Scapulohumeral rhythm using the inverse dynamics model to compensate for the dynamics of the robot (including gravity force). According to the authors, to provide the therapeutic movement the main contributor of torque was the effect of gravity, since the movement has low velocity, therefore inertial forces can be ignored. Firstly, the feed-forward torque with zero-torque compensated the majority of the weight of the robot against gravity was confirmed in every configuration. To constrain the shoulder mechanism, the coupling torque was added to follow the upper arm link with the given angular ratio. The trajectory of the shoulder mechanism with respect to the upper arm link angles were tracked throughout the elevation with a nearly zero coupling torque. On the contrary, when the force is applied to the shoulder girdle mechanism while the operator elevates (which mimics an abnormal scapulohumeral rhythm), the shoulder mechanism exerts a gentle force to recover a normal coordinated angle safely. \\
In \cite{ARMin} the authors have implemented a feedforward model based arm weight compensation using the rehabilitation Robot ARMin on the upper and lower arm. They have performed an evaluation of their requirements and according to that, their exoskeleton should have Freedom of Movement, No Additional Disturbances, Scalability, and Applicability to other systems. Afterward, the experimental results were verified using EMG measurements. They installed their EMG electrodes on 6 different muscles and performed EMG measurements in 4 different positions on 3 subjects. They found out that weight compensation reduces the effort of the subject wearing the weight compensated robot, by an average of 26\% and across the whole workspace. Their feedforward model-based weight compensation method took the torque data from the robot as the input and performed the suitable actuation. Their method is quite similar to our approach, the only difference is that their feedforward control system is based on the torque inputs, which do the weight compensation afterward. \\
The manufacturers and authors of HomeRehab Robotic Device \cite{Inaki1} and \cite{Inaki2} have implemented analytical and machine learning methods to perform gravity compensation. Their 3 DoF robot which consists of closed chain kinematics even though not an actual wearable exoskeleton, but it tries to accomplish the task of rehabilitation of stroke patients as well as perform gravity compensation in 2 dimensions or 3 dimensions \cite{Inaki1}. Their control system is based on feedback from an external, low-cost force sensor in the end effector. Afterward, they devised analytical equations for gravity compensation and compared them with their PID controller and several machine learning algorithms. According to their results, it was difficult to ascertain which performed better ML methods or Analytical equations for the determination of gravity compensation torques. Analytical Methods can calculate the gravity compensation torque even outside the workspace whereas, ML methods require only training data to learn the compensating torques, and they cannot perform outside the workspace they are trained for. This drawback is more like an overfitting problem, but still can be solved by using deep learning methods and more training data.\\
In the current work, we developed a 4 DoF exoskeleton, which performed active gravity compensation in 3 DoF, with a 3-dimensional workspace by implementing a feedforward control system using only the positional inputs from the motor’s internal sensors for controlling the output torques. In the above-mentioned works, they have either used feedback controllers (PD, PID, etc.) with external sensors or used feedforward controllers using internal torque sensors for performing active or passive gravity compensation. \\
Even though in our case, using internal positional sensors and a feedforward controller made the modeling equations of the system more complex, it also contributed to making the system more lightweight and responsive. 

\section{Theory}
\label{Theory}

To determine the joint torques and forces necessary to generate the desired joint positions' corresponding acceleration, we employ the concept of inverse dynamics. In this study, we have adopted the Newton-Euler Inverse Dynamics approach, which relies on the forces exerted on each individual link. Its recursive computational structure, particularly its treatment of rotational dynamics, makes it a suitable choice \cite{Langrangian}.


When an $n$-DoF rigid serial link manipulator experiences no external forces, the Newton-Euler Inverse Dynamics equation can be formulated as shown below:

\begin{equation}
\tau = M(\theta)\ddot \theta + C(\theta,\dot \theta) + F(\dot \theta) + G(\theta) + J(\theta)^T W
\label{Eq1}
\end{equation}

In this equation:
\begin{itemize}
    \item $\theta$, $\dot \theta$, and $\ddot \theta$ represent $N \times 1$ vectors representing joint positions, velocities, and accelerations, respectively.
    \item $M(\theta)$ represents an $N \times N$ matrix representing the symmetric joint space inertia matrix.
    \item $C(\theta,\dot \theta)$ represents an $N \times N$ matrix representing the symmetric Centrifugal and Coriolis forces.
\item   $F(\dot \theta)$ represents an $N \times 1$ vector representing joint frictions.
\item $G(\theta)$ represents an $N \times 1$ vector representing the gravitational influence on each joint.
\item $J(\theta)$ represents a $6 \times N$ matrix, which is the Jacobian Matrix transforming the force at the End-Effector (located at the $N^{th}$ link) into joint torques.
\item $W$ is $6 \times 1$ a vector which consists of $[F_x F_y F_z  M_x M_y M_z]^T$ where $F$ and $M$ represent forces and moment respectively on the $N^{th}$ link of the End-Effector.
\item $\tau$ is the $N \times 1$ vector consisting of torque at each joint
\end{itemize}

In Equation \ref{Eq1}, we should take into consideration the different variables of that equation in the context of the exoskeleton.
Based on the calculations, experimentation, and literature review, it is evident that the inertia matrix,
$M(\theta)\ddot \theta $, which depends upon the acceleration, length, mass, and radius of the human and the robot has a very negligible impact on the overall gravity compensation as all of those variables are quite small. This can be confirmed by \cite{Inertia}, where the authors performed the mathematical analysis of the moment of inertia of the human arm at a fixed position while considering the human arm as the frustum of a cone (truncated cone). Similarly, the numeric values of inertia generated by the human body are presented by\cite{Body_Inertia} which also indicates the low values by the human arm. \\
Similarly, \cite{H.Flash}, studied the effects of velocity torques on the human arm at different velocities and came to the conclusion that when the movement of the human arm is slow, gravity plays a major role in dynamics whereas, at higher velocities, torque related to velocity takes over. Our work is aimed toward the rehabilitation of stroke patients using the exoskeleton, therefore, its overall velocity is kept low. We also noticed the same thing as the influence of gravity is much higher than the effect of inertia or centrifugal force since the exoskeleton design is lightweight and the subject cannot move his/her arm at a higher velocity. \\
The Coriolis forces and centrifugal forces represented by  $C(\theta,\dot \theta)$  do not influence the robot in any meaningful way. The viscous and coulomb frictions $F(\dot \theta)$ exist in the system stemming from the internal frictions of the motors, but their overall effect does not influence the gravity compensation. 
$J(\theta)^T W$ is due to the end-effector, which the current exoskeleton does not have. 
The effect of gravity as represented by $G(\theta)$ in Equation \ref{Eq1} has the most influence on the exoskeleton dynamics as each pose has a different orientation of the joints, therefore, the effect of gravity is different which is calculated in Equations \ref{eq:EQ2} to \ref{eq:EQ5} and Equations \ref{eq:EQ1m} to \ref{eq:EQ4m} with gyroscopic stabilization. 

\section{Design of Exoskeleton}
\begin{figure}[htp]
    \centering
    \includegraphics[width=5cm]{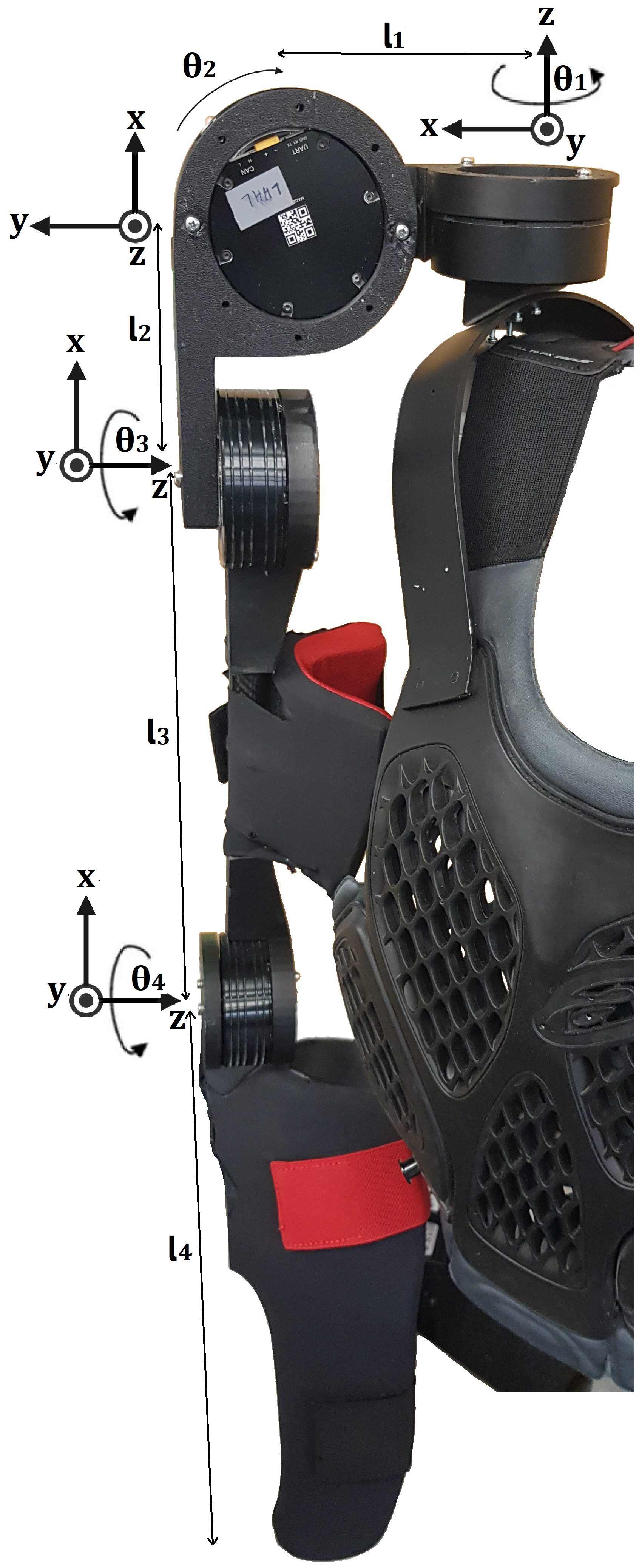}
    \caption{Schematic Diagram of Exoskeleton}
    \label{fig:1}
\end{figure}

\begin{figure}[htp]
    \centering
    \includegraphics[width=5cm]{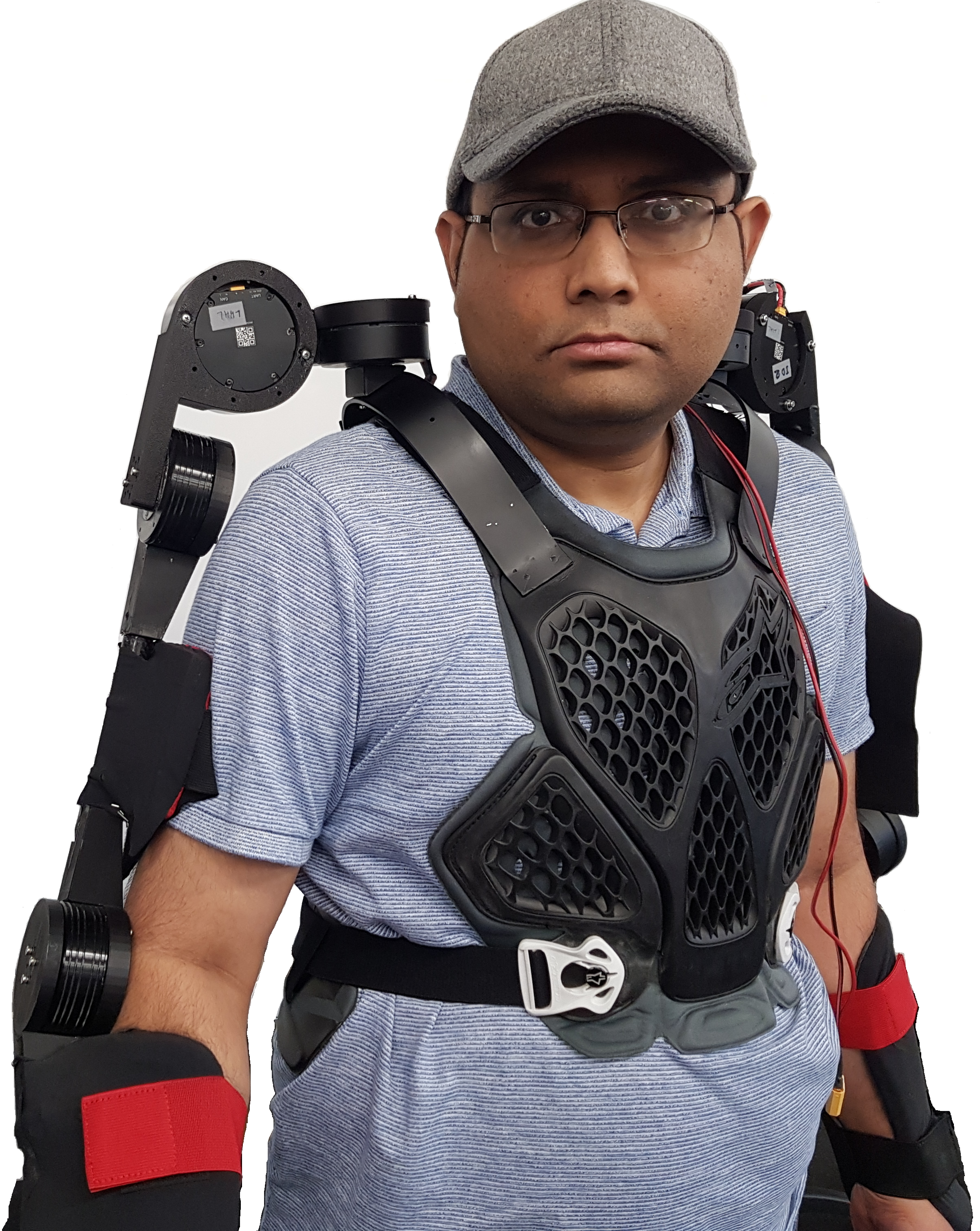}
    \caption{Exoskeleton with the wearer}
    \label{fig:2}
\end{figure}

In this study, an upper-body exoskeleton is developed, which consists of 4 high power-to-weight ratio Brushless DC (BLDC) Motors on each side. The shoulder joints are represented by 3 Motors mounted on the top side of the Exoskeleton, and the fourth motor represents the elbow joint in Figures \ref{fig:1} and \ref{fig:2}. Each motor has position, velocity, and current sensors. The overall assembly including both the left and right sides of the exoskeleton, the wearing assembly, power pack, electric and CAN bus connections, circuitry, and controllers weighs around 8 kg and is highly mobile and ergonomic as weight is quite distributed. The workspace of the exoskeleton is 3-dimensional, therefore it is capable of movement and can perform gravity compensation in multiple planes and directions. The exoskeleton is connected to  the user’s arms with cuffs at the bicep and forearm so that the user can control the robotic arm as well as the motion of the shoulder joints.

\section{Methodology}

The primary objectives revolved around minimizing both weight and hardware complexity while ensuring a highly responsive performance to eliminate any perceptible delay in gravity compensation during motion for the exoskeleton user. Consequently, a Feedforward Controller was devised due to its ability to offer the desired advantages. This controller derives gravity-compensating torques for each joint by utilizing input data from the internal position sensor of each motor.

One of the key advantages of implementing the Feedforward Controller is its obviation of the need for an additional external sensor in the exoskeleton hardware, such as a wearable torque sensor. This decision not only keeps costs and complexity at a minimum but also enhances ergonomic considerations.

Furthermore, \cite{H.Flash} conducted a comparison between open-loop (feedforward) and closed-loop (feedback) controllers, highlighting that open-loop controllers, which precisely compute joint torques analytically, exhibit superior reactivity when compared to their feedback (closed-loop) counterparts. This equates to reduced computation time.

Additionally, since the system's dynamics were meticulously taken into account during the derivation of the system's modeling equations in the exoskeleton design, it effectively overcame the typical limitation of feedforward systems, which necessitate careful consideration of all system parameters to be controlled.

\begin{figure}[htp]
    \centering
    \includegraphics[width=6cm]{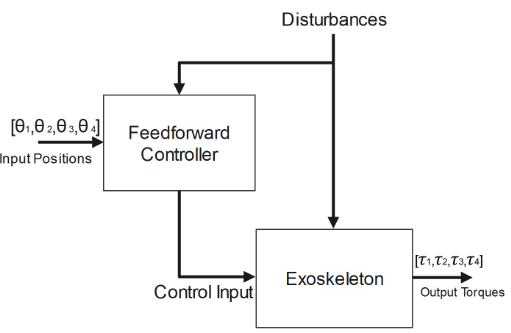}
    \caption{Feedforward Scheme for the control of Exoskeleton for Equations \ref{eq:EQ2} to \ref{eq:EQ5} }
    \label{fig:3}
\end{figure}

\begin{figure}[htp]
    \centering
    \includegraphics[width=8cm]{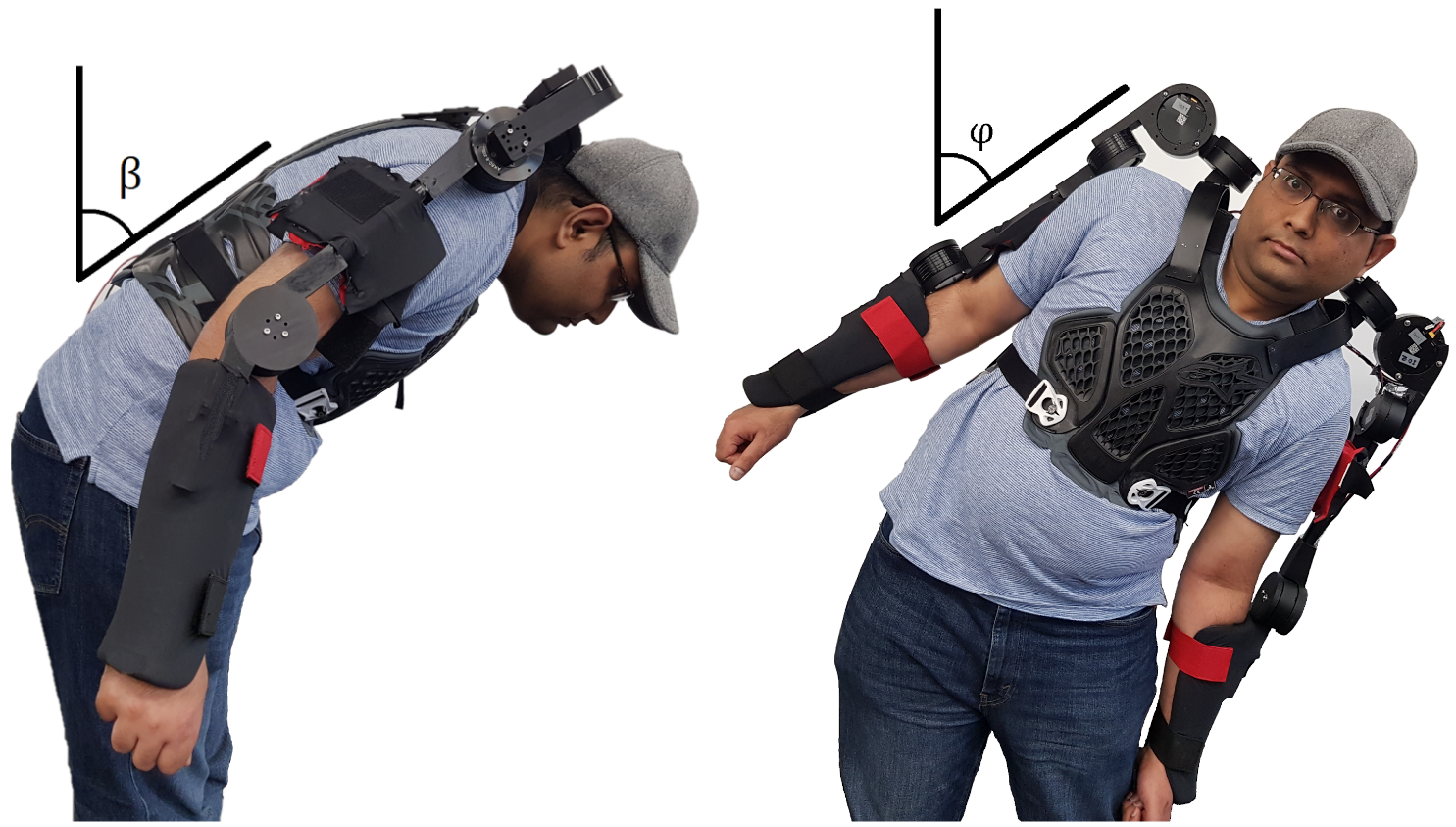}
    \caption{Bowing and Tilting, Represented by $\beta$ and $\phi$  respectively}
    \label{fig:4}
\end{figure}

The inverse dynamics equations employed in this scenario are outlined below. When the subject is stationary, particularly when seated or immobile, the associated torque for each motor joint is determined as follows. These equations are specifically applicable when the robot's design adheres to the DH Parameters, as detailed in Table \ref{table_example}.
\begin{align}
     \tau_{1} &= c_1g[d_1(m_1+m_2+m_3+m_4) \nonumber \\
     &+ l_2(m_2+m_3+m_4)\sin(\theta_2) \nonumber \\
     &+ l_3(m_3+m_4)\cos(\theta_3) \nonumber \\
     &+ l_4m_4\cos(\theta_3+\theta_4)]\sin(\theta_1)\label{eq:EQ2}\\[2ex]
    \tau_{2} &= c_2g [d_2(m_2+m_3+m_4)+ l_3(m_3+m_4) \nonumber \\
    &\cdot \cos(\theta_3) + l_4m_4\cos(\theta_3+\theta_4)]\sin(\theta_2)\label{eq:EQ3}\\[2ex]
     \tau_{3} &= c_3g[d_3(m_3+m_4)\sin(\theta_3) \nonumber \\
     &+ l_4(m_4)\sin(\theta_3+\theta_4)]\cos(\theta_2)\label{eq:EQ4}\\[2ex]
     \tau_{4} & =  c_4 g d_4(m_4) \sin(\theta_3+\theta_4)\cos(\theta_2)
        \label{eq:EQ5}
\end{align}
Where  $l_n$ is the length of the corresponding link,  $d_n$ is the center of gravity, g = 9.8 $m/s^2$ is the acceleration due to gravity,
 $m_n,\theta_n,\tau_n$, are the mass, angular position, and torque respectively at the $n$-th joint\\
 $c_n$  is the constant to adjust/vary the value of torque for the $n$-th joint (by default, it should be kept at 1).\\
$\theta_n$ the angle motor joint makes against the gravity vector.\\
$ \theta_n  = [ \forall \in \theta_1,-\pi/2,0,0]$ corresponds to the straight downward position from joint 2 (of minimum potential energy).
$l_1,l_2,l_3,l_4$ represent the corresponding lengths of the links as shown in Figure \ref{fig:1}.\\
Joint 1, if the wearer is not moving or by default, its axis is parallel with the gravity vector, there is no need for gravity compensation in this case as $sin\theta_1=0$ (i.e. its z-axis is parallel to the gravity vector) therefore, $\tau_1=0$ in Equation \ref{eq:EQ2}. \\
Joint 2 is orthogonal to the subsequent joints 3 and 4. So initially, when Joints 3 and 4 are at the default position(i.e.$(\theta_3, \theta_4)= (0,0))$, it works as a simple planar arm with a combined mass and lengths of all subsequent joints. But when Joints 3 and 4 are at some non-zero angle, the combined torques affecting Joint 2 should be less since the length of the moment arm is shorter than the length at the default position.\\
If the subject is wearing the exoskeleton and is mobile, then the stabilization must be provided using the outputs of the gyroscopes mounted on the flat surface near the shoulder. The angles of pitch (bowing) and yaw (sideways movement) of the human body, as shown in Figure \ref{fig:4}, must be accounted for which are represented by $\beta$ and $\phi$ respectively. The following set of equations (Equations \ref{eq:EQ1m} to \ref{eq:EQ4m}), considering that their default angles $(\beta,\phi)=(0,0)$ were built on top of the above-mentioned set of Equations \ref{eq:EQ2} to \ref{eq:EQ5} with input from gyroscopic sensors considered. In this case, $\tau_1$ should be considered non-zero as $\theta_1$ may or may not be aligned with the gravity vector and is represented by $\tau_{1m}$.
\begin{align}
     \tau_{1m} &= c_1g[d_1(m_1+m_2+m_3+m_4) \nonumber \\
     &+ l_2(m_2+m_3+m_4)\sin(\theta_2+\phi) \nonumber \\
     &+ l_3(m_3+m_4)\cos(\theta_3+\beta) \nonumber \\
     &+ l_4m_4\cos(\theta_3+\theta_4+\beta)]\sin(\theta_1)\cos(\phi)\cos(\beta) \label{eq:EQ1m} \\[2ex]
    \tau_{2m} &= c_2g [d_2(m_2+m_3+m_4) \nonumber \\
    &+ l_3(m_3+m_4)\cos(\theta_3+\beta) + l_4m_4\cos(\theta_3+\theta_4+\beta) ] \nonumber \\
     &\cdot \sin(\theta_2+\phi)\cos(\theta_1)\cos(\phi)\cos(\beta)\label{eq:EQ2m}\\[2ex]    
     \tau_{3m} &= c_3g[d_3(m_3+m_4)\sin(\theta_3+\beta) \nonumber \\
     &+ l_4(m_4)\sin(\theta_3+\theta_4+\beta)]\cos(\theta_2+\phi) \nonumber \\
     &\cdot \cos(\theta_1)\cos(\phi)\cos(\beta) \label{eq:EQ3m}     \\[2ex] 
     \tau_{4m} & = c_4gd_4(m_4)\sin(\theta_3+\theta_4+\beta) \nonumber \\
      & \cdot \cos(\theta_2+\phi) \cos(\theta_1)\cos(\phi)\cos(\beta)     \label{eq:EQ4m}
\end{align}
Where $\tau_{nm}$ is the torque at the respective $n$-th joint.
\section{Implementation of Feedforward Controller}
In this section, we shall discuss the implementation of the control equations in Simulation and on actual hardware. 
\subsection{Simulation and Data-Driven Model for Gravity Compensation Equation}
\label{Definition of Robot}
The Feedforward Controller was first tested on an equivalent representation of the exoskeleton on Matlab and Simulink using the Robotics Toolbox by Peter Corke \cite{PETER_corke}.  It comes with a wide range of functions that enable users to effortlessly visualize and regulate the motion and dynamics of individual serial-linked robotic manipulators. To create the linkages, the functions {\tt Link(dh, options)} and {\tt SerialLink(L)}  were utilized, first to construct the individual links and then to connect them. The Robotic Toolbox utilizes the Denavit-Hartenberg(DH) parameters to assign the robotic link configuration and properties $[a_n,\alpha_n,d_n,\theta_n]$. For the designed Exoskeleton, the DH parameter table is shown in Table \ref{table_example}.  The dynamic parameters of the exoskeleton were defined afterward, including, link mass, viscous friction, coulomb friction, motor inertia, inertia matrix, gravitational vector, the center of gravity, and gear ratio. The offset of $-\pi/2$ rad is provided to joint 2 such that by default the exoskeleton hangs downward like a human arm from joint 2 onwards.  
\begin{table}[h]	
  \renewcommand{\arraystretch}{1.3}
	\begin{center}
		\begin{tabular}{|c|c|c|c|c|c|}
			\hline
			 & $a(m)$ & $\alpha(rad)$ & $d(m)$ & $\theta(rad)$ & $offset(rad)$ \\
			\hline \hline
			Joint 1 & 0.05 & $-\pi/2$ & 0 & $\theta_1$ & 0\\
			\hline
   Joint 2 & 0.13 & $\pi/2$ & 0 & $\theta_2$ & $-\pi/2$\\
			\hline
   Joint 3 & 0.3 & 0 & 0 & $\theta_3 $ & 0\\
			\hline
   Joint 4 & 0.3 & 0 & 0 & $\theta_4$ & 0\\
			\hline
		\end{tabular} 
  \vspace{2ex}
        \caption{DH Parameters of the Exoskeleton}
        \label{table_example}
	\end{center}
\end{table}

After defining the robot, the {\tt gravload} function of Robotics Toolbox was used to simulate the effect of gravity from any starting position of the robot. The {\tt gravload} function is based on RNE (Recursive Newton Euler) which calculates the inverse dynamics of the system under consideration. The output of the {\tt gravload} function was used to compare the accuracy of the analytical function and Adaptive Network-based Fuzzy Inference System (ANFIS) controller in the simulation. The ANFIS controllers are based on the combination of Artificial Neural Networks(ANN) and Fuzzy Inference System(FIS). The ANN part does the classification of the input, whereas the FIS part of ANFIS decides the output based on linguistic models. \\
Then the model is given the command to move within the workspace, while the {\tt gravload} function and analytical equations (Equations \ref{eq:EQ3} to \ref{eq:EQ5}) calculate the amount of torque exerted on every joint at different joint angles on certain intervals under the influence of gravity. Using the torque data generated by {\tt gravload} function, at different joint angles, an ANFIS controller model was trained.  Afterward, the comparison was made between the output of the implemented {\tt gravload} function, analytical equations, and ANFIS controller. Based on the comparison, it was found that the analytical equations performed much better than the ANFIS controller, almost similar to the benchmark {\tt gravload} function. The resulting Root Mean Square Error (RMSE) is given in Section \ref{Result of Sim}. 

\subsection{Implementation of Analytical Equation on Hardware}


Following the verification of analytical equations through simulation trials, they were subsequently put into practice on the exoskeleton. A comprehensive hardware description can be found in Section \ref{Appendix}. The controller setup involved utilizing an Arduino Uno equipped with a CAN bus shield, responsible for managing four BLDC Motors on each side through the CAN bus communication protocol. For an evaluation of the exoskeleton's stability during the execution of gravity compensation, please refer to Section \ref{Result of Hardware}.

\section{Results}
\label{Results}
This section is divided into the simulation and hardware parts. In the simulation part, both the Analytical Equations for calculating the torque corresponding to each joint as stated in Equations \ref{eq:EQ3} to \ref{eq:EQ5}, and the output generated by ANFIS controller were compared with the {\tt gravload} function (mentioned in Section \ref{Definition of Robot}) of Robotics toolbox. Afterward, in the hardware part, the result of the stability analysis while performing the active gravity compensation at multiple joint positions of the exoskeleton is discussed. 
\subsection{Results from Simulation}
\label{Result of Sim}
The error size between the joint torques calculated by inverse dynamics function of the Robotics Toolbox ({\tt gravload}) on the open kinematic chain (Exoskeleton) when compared with Analytical Equations \ref{eq:EQ3} to \ref{eq:EQ5} and ANFIS respectively for each joint for 1053 different positions spanning the entire workspace of the Exoskeleton. The Root Mean Squared Error (RMSE) of the analytical function and ANFIS were calculated for their accuracy, and it was inferred that the analytical function has negligible RMSE (for Joint 2, Joint 3, and Joint 4 as $6.06e^{-16}$, $4.06e^{-16}$, $8.03e^{-17}$  respectively), whereas ANFIS Controller, RMSE for  Joint 2, Joint 3 and Joint 4 was $1.71e^{-3}$, $1.32e^{-3}$, $3.15e^{-4}$respectively which is significantly higher than analytical equations. ANFIS takes around 380 seconds on average to train to predict one joint torque from the simulation data on AMD Ryzen 5950HX Processor, 32 GB RAM, and Nvidia RTX3080 GPU on Matlab. 
\subsection{Results from Hardware}
\label{Result of Hardware}
\begin{figure}[htp]
    \centering
    \includegraphics[width=.9\columnwidth]{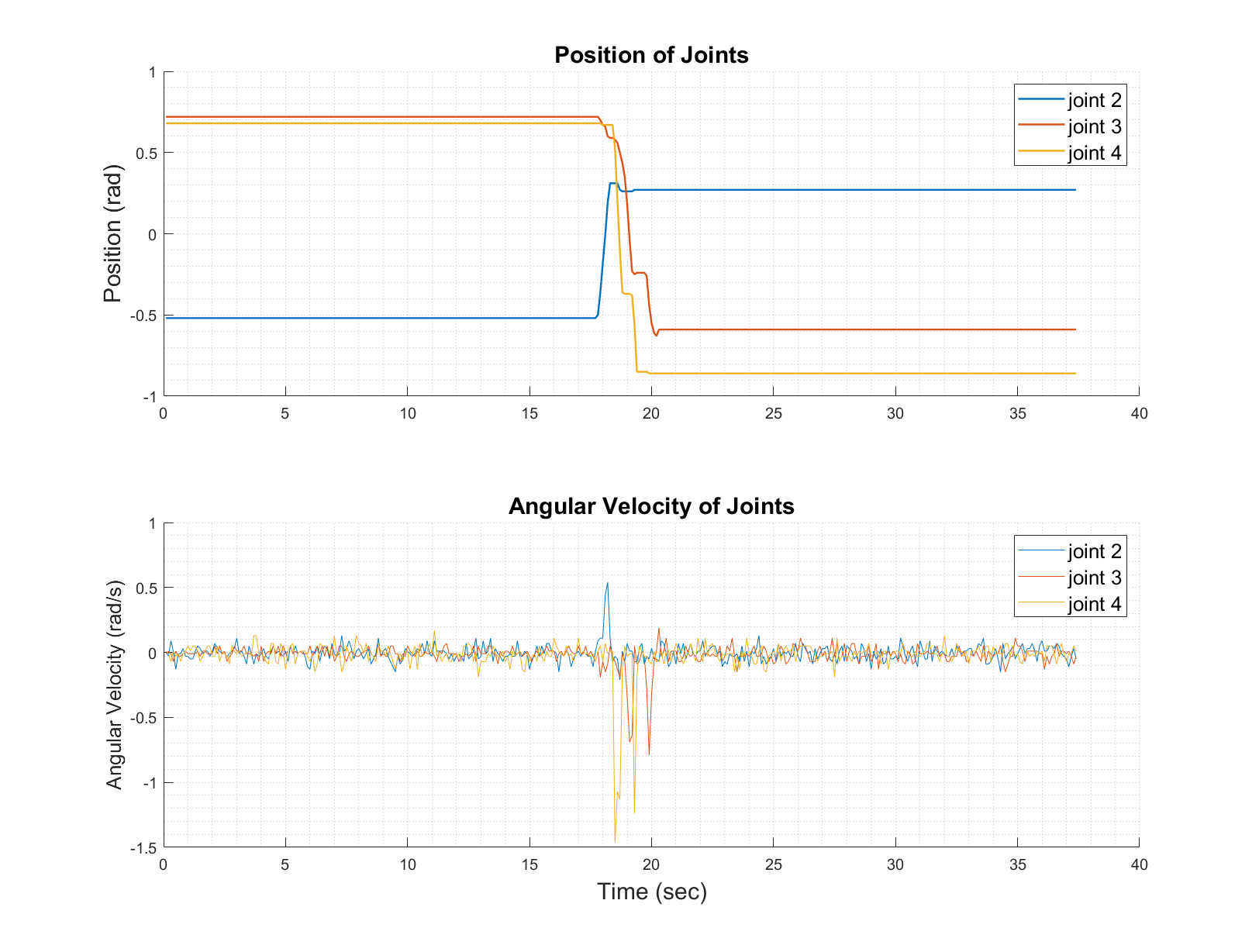}
    \caption{Stability analysis during the active gravity compensation of the Exoskeleton for two different positions}
    \label{fig:one_pos}
\end{figure}

\begin{figure}[htp]
    \centering
    \includegraphics[width=.9\columnwidth]{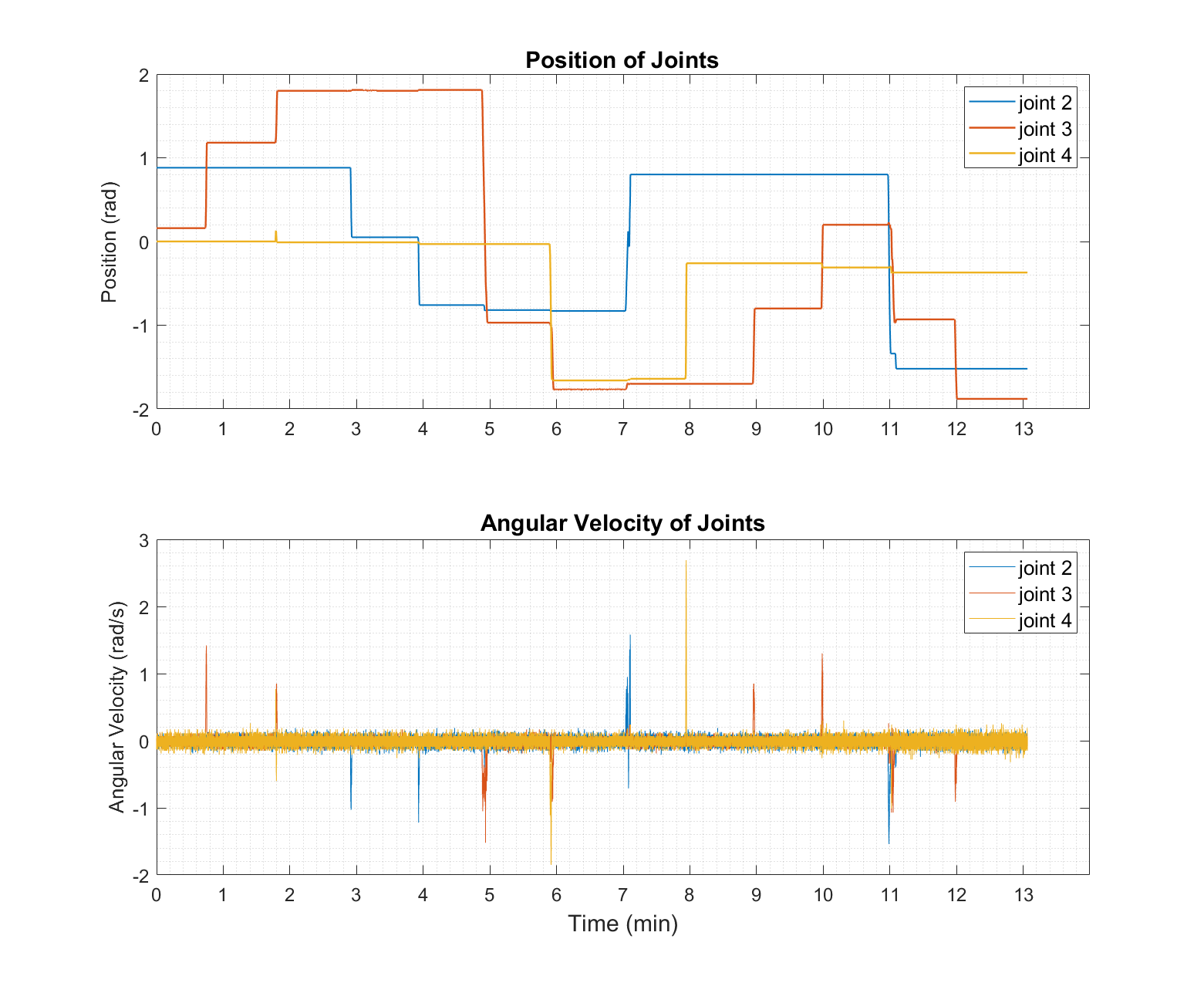}
    \caption{Stability analysis during the active gravity compensation of the Exoskeleton for multiple positions}
    \label{fig:13_pos}

\end{figure}

Gravity compensation means that the robot should hold its particular position without succumbing to the influence of gravity and collapsing. Therefore, this can be ascertained that there is no variation in position when active gravity compensation is implemented on it. In the first subplot of Figure \ref{fig:one_pos}, the positions of three joints of the robot (Joints 2, 3, and 4) that are not parallel with the gravity vector are shown. They will fall to their lowest potential energy state without gravity compensation. However, when gravity compensation is applied, they stay at their respective positions, indicating that the gravity's overall effect is nullified. \\
The middle section of Figure \ref{fig:one_pos}  highlights a disturbance in positions and velocities that corresponds to the deliberate manipulation of every joint by the user within the 17-second to 20-second timeframe. In which, Joint 2 was elevated whereas joints 3 and 4 were moved downwards (as seen from Positional and Velocity subgraphs). Following this movement, all the joints maintained their respective positions and their velocities were around zero. \\
In Figure \ref{fig:13_pos}, the stability analysis was performed for multiple joint orientations of the exoskeleton. Each position was held for about a minute, and then the joints were rotated to a new position and were tested for gravity compensation for that particular position before transitioning to a new position. In total, 13 different positions are shown here for reference. After moving to a new position, the joints held their position as there was no change in position, implying zero overall velocity. The video of hardware performing the gravity compensation at One-G is available at \cite{urlGrav}.
\section{Discussion}
This study focuses on the conceptualization and control system of a lightweight upper-body exoskeleton. The hardware was kept simple to reduce the weight such as the use of a high power-to-weight ratio BLDC Motors with built-in driver circuitry in the joints, lightweight carbon fiber for the link construction, and a decision to forgo external sensors due to the use of a Feedforward controller. \\
The analytical equations were opted over the data-driven approach of the ANFIS controller due to their lower accuracy. There are shortcomings of ANFIS controllers such as the curse of dimensionality, which means that if the number of features increases then it requires an exponential amount of time to train with time complexity around the order of ($O^2$) \cite{anfis}.  Also, ANFIS controllers are Multiple Input Single Output (MISO) systems, which means that we have to train one ANFIS for every joint that takes the positions ($\theta_1, \theta_2, \theta_3, \theta_4$)  of all 4 Joints and calculates the torque ($\tau_n$)  for that joint. If we add the stabilization data from the gyroscope, then it shall have more input features($\theta_1, \theta_2, \theta_3, \theta_4,\beta,\phi$) and would be much slower to train. Therefore, in our case, we have to put 4 ANFIS controllers into our microcontroller (each outputting the corresponding torque for every motor) which requires a bit more processing than the system of the analytical equations. Since ANFIS needs the positional data and corresponding torque of the motors to train, data generation from actual hardware was a time-consuming step but in simulation, the data, and feature generation was easier to do as compared to the hardware of the exoskeleton. The analytical function worked perfectly in Simulation as it generated almost negligible RMSE and the robot held its position under the influence of gravity even with minimal friction (emanating from the motors) the actual hardware was totally stable as it did not drift from its given position for any given time-frame.

\section{Conclusion and Future Works}
In this paper, we have successfully demonstrated the feasibility of gravity compensation on an upper body exoskeleton or similar robot using the Feedforward Analytical Function based on Newton-Euler Inverse Dynamic Equations. Through rigorous testing on both simulations and a physical hardware prototype of the exoskeleton, the controller based on these equations exhibited flawless performance. Notably, in comparison to the data-driven ANFIS controller, the analytical solution proved to be more accurate in the simulation, leading us to disregard the ANFIS controller for the exoskeleton hardware.\\
Looking ahead, if flexible or elastic links are employed in the system, the analytical solution may become impractical due to uncertainties introduced by elasticity. In such cases, data-driven approaches like ANFIS could be considered instead.\\
Currently, the exoskeleton prototype effectively compensates for its own weight (One-G). However, future iterations will require more powerful motors to enable compensation for the wearer's arm weight as well as the robot's weight (Zero-G). The gravity compensation equations will remain unchanged, with adjustments made only to the corresponding mass constant to accommodate the varying arm mass.
While the exoskeleton controller currently provides gravity compensation, we propose the development of an additional Feedback control loop to complement the existing feed-forward loop. This new loop will be designed to control the exoskeleton's position and velocity using external sensors like IMEI, EMG, EEG sensors, or VR glasses. The objective is to create a cascade control system and determine if stroke patients, while wearing the exoskeleton, can effectively control it using external signals apart from their arm muscles. This would open up possibilities for rehabilitation or other tasks.\\
In conclusion, our research showcases the efficacy of the Feedforward Analytical Function in gravity compensation for an upper body exoskeleton and outlines a promising direction for further enhancing its control capabilities through the integration of external sensors in a cascade control system.

\section{Appendix}
\label{Appendix}
\subsection{Hardware Specifications}
\subsubsection{Motors}
On each side of the exoskeleton, there are four BLDC Motors, three of them are AK-60-6 (Joints 1,3 and 4) and one AK-80-9 Motor (Joint 2) from T-motors.\\
The AK-60-6 is a BLDC motor designed for use in robotics applications. It is a high-performance motor with a voltage rating of 24 volts, a rated current of 7.4 amps, a peak torque of 9 Nm, and a weight of 315 grams. T-Motor AK-80-9 is a larger derivative of the AK-60-6 and has a voltage rating of 48 volts, a rated current of 10.3 amps, a peak torque of 18 Nm, and a weight of 485 grams.  They have a compact size, inbuilt driver circuitry, low weight, reliability, and a high-to-weight ratio making them suitable for use in small robots, robotic arms, or other mechanisms that require precise and high-force movement such as upper-body exoskeleton. 
\subsubsection{Links and Brackets}
Lightweight Carbon Fiber tubes are used in the links and the brackets of the motor mounts. The 3D-printed Motor mounts are built for 190° (approx. 3.3 rad) movements. They are durable,  have a high strength-to-weight ratio, corrosion resistance, and low thermal expansion. 
\subsubsection{Controller, Communication, and Power Supply}
The control system was implemented on an Arduino Uno which is based on the ATmega328P microcontroller. The programming was done using Arduino Integrated Development Environment(IDE) which supports a user-friendly simplified version of the C++ programming language.  As the motors communicated via CAN bus protocol, the CAN shield was used on top of the microcontroller. \\
The whole setup was powered by a TATTU Smart Lipo Battery 22,2V 6 Cells, 222Wh which weighs around 1.4 kg. It is compact, lightweight, has high energy density, high output, and long lifespan.

 \bibliographystyle{unsrt}
\bibliography{relatedWork}
\end{document}